\documentclass[10pt,twocolumn,letterpaper]{article}

\usepackage{iccv}
\usepackage{times}
\usepackage{epsfig}
\usepackage{graphicx}
\usepackage{amsmath}
\usepackage{amssymb}

\usepackage{multirow}
\usepackage{array}
\usepackage{dcolumn}

\usepackage{graphicx}
\usepackage{subfig}
\usepackage{comment}
\usepackage[breaklinks=true,bookmarks=false]{hyperref}

\iccvfinalcopy % *** Uncomment this line for the final submission

\def\iccvPaperID{2111} % *** Enter the ICCV Paper ID here

\ificcvfinal\fi
\begin{document}

%%%%%%%%% TITLE
\title{A Multilayer-Based Framework for Online Background Subtraction \\with Freely Moving Cameras}
\author{Yizhe Zhu \quad\quad\quad\quad\quad\quad\quad\quad\quad\quad Ahmed Elgammal\\
	Department of Computer Science, Rutgers University, Piscataway, NJ 08854\\
	{\tt\small yizhe.zhu@rutgers.edu} \quad\quad\quad\quad\quad\quad
	{\tt\small elgammal@cs.rutgers.edu} 
}

\begin{comment}

\author{Yizhe Zhu, \quad\quad\quad\quad\quad \quad\quad  Ahmed Elgammal \\
	yizhe.zhu@rutgers.edu,  \quad\quad\quad elgammal@cs.rutgers.edu \\
	Rutgers University, Department of Computer Science
}
	content...
	\end{comment}
\begin{comment}

\author{First Author\\
Institution1\\
Institution1 address\\
{\tt\small firstauthor@i1.org}
% For a paper whose authors are all at the same institution,
% omit the following lines up until the closing ``}''.
% Additional authors and addresses can be added with ``\and'',
% just like the second author.
% To save space, use either the email address or home page, not both
\and
Second Author\\
Institution2\\
First line of institution2 address\\
{\tt\small secondauthor@i2.org}
}
\end{comment}
\maketitle
%\thispagestyle{empty}

%%%%%%%%% ABSTRACT
\begin{abstract}
The exponentially increasing use of moving platforms for video capture introduces the urgent need to develop the general background subtraction algorithms with the capability to deal with the moving background. In this paper, we propose a multilayer-based framework for online background subtraction for videos captured by moving cameras. Unlike the previous treatments of the problem, the proposed method is not restricted to binary segmentation of background and foreground, but formulates it as a multi-label segmentation problem by modeling multiple foreground objects in different layers when they appear simultaneously in the scene. We assign an independent processing layer to each foreground object, as well as the background, where both motion and appearance models are estimated, and a probability map is inferred using a Bayesian filtering framework. Finally, Multi-label Graph-cut on Markov Random Field is employed to perform pixel-wise labeling. Extensive evaluation results show that the proposed method outperforms state-of-the-art methods on challenging video sequences.
\end{abstract}

%%%%%%%%% BODY TEXT

%===========================================================
\section{Introduction}
The identification of regions of interest is typically the critical preprocessing step for various high-level computer vision applications, including event detection, video surveillance, human motion analysis, etc. Background subtraction is a widely-used technique to perform pixel-wise segmentation of foreground regions out of background scenes. Unlike foreground object detection algorithms, background subtraction methods typically produce much more accurate segmentation of foreground regions rather than merely detection bounding boxes, without the need to train individual object detectors.  
A great number of various traditional background subtraction methods and algorithms have been proposed \cite{Stauffer1999,Elgammal2002,DarShyangLee2005,Sheikh2005,Han2012,Ren2013}. Most of these methods focused on modeling background under the assumption that the camera is stationary. However, more and more videos are captured from moving platforms, such as camera phones, and cameras mounted on ground vehicles, robots, ariel drones, etc. Traditional background subtraction algorithms are no longer applicable for such videos captured from a non-stationary platform \cite{Elgammal2014}. 
The exponentially increasing use of moving platforms for video capture introduces a high demand for the development of general background subtraction algorithms that are not only as effective as traditional background subtraction but also applicable to moving-camera videos.

%In contrast to those captured by stationary cameras, the pixels of the background scene in moving-camera videos no longer maintain their positions in consecutive frames, naturally making background subtraction more challenging. 
Similar to most video segmentation methods, a few works \cite{Brox2014,Cui2012} resort to processing the whole video offline. Offline methods can typically produce good results on short sequences since the information in latter frames can significantly benefit the segmentation in the earlier frames. However, since it needs to store and process the information over the whole video, the memory and computational cost increase exponentially as the number of frames to process grows~\cite{Elgammal2014}. Additionally, in various cases, such as video surveillance and security monitoring, videos need to be analyzed as they being streamed in real time, where an efficient online background subtraction method is greatly in demand. 

The key to handling long sequences in an online way is to learn and maintain models for the background and foreground layers. Such models accumulate and update the evidence over a large number of frames and also supply valuable knowledge foundation to high-level vision tasks. Recently, a few online background subtraction methods with moving cameras have been proposed \cite{Sheikh2009,han2011,Ali2012,Lim2014,Zamalieva2014}. Most methods formulate it as a binary segmentation problem with the assumption of only one foreground object, naturally resulting in bad segmentation when multiple moving objects appear in the scene. Especially in the case where objects go across each other, motion estimation for objects suffers great confusion, which further degrades the performance of background subtraction.

To remedy this drawback, we propose a general multilayer-based framework with the capability of handling multiple foreground objects in the scene. 
The objects can be automatically detected based on motion inconsistency and an independent processing layer is assigned to every foreground object as well as the background. In each layer, we would like the same process to be preformed concurrently inside the ``processing block", which takes the accumulated information and the new evidence of each layer as input, and outputs a probability map indicating the confidence of pixels belonging to each layer. In this paper, we elaborately design such a ``processing block" with three steps as follows. 
(a) Motion model estimation is first performed based on Gaussian Belief Propagation \cite{GBP2008} with motion vectors of corresponding trajectories as the evidence. (b) The appearance model and the prior probability are predicted by propagating the previous one with the estimated motion model.  (c) Given the current frame as new evidence, Kernel Density Estimation \cite{Elgammal2002} is employed to infer the probability map as the output. Finally, based on the collection of probability maps produced by each ``processing block", the pixel-wise segmentation for the current frame is generated by Multi-label Graph-cut.

Besides, since it is the first work to tackle multilabel background subtraction problem in moving camera scenarios, to our best knowledge, we also design a methodology to evaluate the performance and show our method outperforms the state-of-the-art methods.

\section{Related Work}

\noindent \textbf{Motion Estimation and Compensation:}
The freely moving camera introduces a movement in the projected background scene, and thus complicates the background subtraction problem. An intuitive idea to tackle such a problem is compensating the camera motion. A few pioneering works resort to estimating a homography \cite{Hartley2004} that characterizes the geometric transformation of background scene between consecutive frames. Typically RANSAC \cite{Fischler1981} and its variants MLESAC \cite{Torr2000} are employed to achieve robust estimation using many matches of feature points. Jin \etal\cite{jin2008} model the scene as a set of planer regions where each background pixel is assumed to belong to one of these regions. Homographies are used to rectify each region to its corresponding planer representation in the model. Zamalieva \etal\cite{Zamalieva2014} leverage geometric information to develop multiple transformation models, and choose one that best describes the relation between consecutive frames.

%The approach computes both homography and fundamental matrices based on estimated dense motion field, and at each frame, one matrix is selected as the geometric transformation that best describes the relation between consecutive frames.

Recently, motion estimation has been widely employed to comprehensively specify the motion for every pixel \cite{ narayana2013, bideau2016s, han2011, Lim2014}. These works \cite{han2011,Lim2014} used optical flow as the evidence. Kwak \etal\cite{han2011} divided the images into small blocks in a grid pattern, and employed nonparametric belief propagation to estimate the motion field based on average optical flow of each block. Its following work \cite{Lim2014} improved the quality of motion estimation by replacing blocks with superpixels as the model unit. %which preserve the visual coherence inside units and avoids motion smoothing across the regions belonging to different objects. 
On the other hand, in \cite{narayana2013, bideau2016s}, optical flow orientations were claimed independent of object depth in the scene, and used to clusters pixels that have similar real-world motion, irrespective of their depth in the scene. However, high dependency on the optical flow makes these methods susceptible to the noise in the estimation of optical flow. In contrast, our method improves motion model estimation by employing Gaussian Belief Propagation \cite{GBP2008} with the motion vectors of sparse feature points as more robust evidence.

\vspace{1.0mm}

\noindent \textbf{Appearance Modeling:}
Traditionally, statistical representations of the background scene have been proposed to estimate spatially extendable background models. Hayman \etal \cite{hayman2003} built a mixture of Gaussian mosaic background model. Ren \etal\cite{ren2003statistical} used motion compensation to predict the position of each pixel in a background map, and model the uncertainty of that prediction by a spacial Gaussian distribution. The construction of image mosaic associated with a traditional mixture Gaussian background model was also claimed to be effective in \cite{Mittal2000,rowe1996}. However, the hyper-parameter required by this parametric model restricts its adaptability and application. On the contrary, we employ nonparametric Kernal Density Estimation method \cite{Elgammal2002} to build models of the appearance of foreground and background regions, making our approach more stable and applicable.

\vspace{1.0mm}
\noindent \textbf{Layered Representation:}
The layered representation, referring to approaches that model the scene as a set of moving layers, has been used for foreground detection \cite{patwardhan2008robust,kim2005background}, motion segmentation \cite{wang1994representing, torr2001integrated, kumar2008learning}.
In \cite{patwardhan2008robust}, the background was modeled as the union of nonparametric layer-models to facilitate detecting the foreground under static or dynamic background. Kim \etal\cite{kim2005background} proposed a layered background model where a long-term background model is used besides several multiple short-term background models. Wang \etal\cite{wang1994representing} used an iterative method to achieve layered-motion segmentation. Torr \etal\cite{torr2001integrated} modeled the layers as planes in 3D and integrating priors in a Bayesian framework. \cite{kumar2008learning} models spatial continuity while representing each layer as composed of a set of segments. A common theme of these layered models is the assumption that the video is available beforehand \cite{Elgammal2014}. Such an assumption prevents the use of such approaches for processing videos from streaming sources. Some dynamic textures methods \cite{chan2008modeling, chan2009layered, mumtaz2014joint} also employed the layered model to tackle the complex dynamic background, but with stationary cameras. To the best of our knowledge, the proposed method is the first layered model applied in the moving camera scenarios. 

The organization of this paper is as follows. The overview of our proposed framework are presented in Section $3$. Section $4$ describes the trajectory labeling. The components inside ``processing block" are described in Section $5\&6$ and the final pixel-wise multi-labeling are presented in section $7$. Finally, Section $8$ illustrates quantitatively and qualitatively experimental results based on two criteria.

\section{Framework Overview}

\begin{figure}%[h!]
	\centering
	\includegraphics[scale = 0.30]{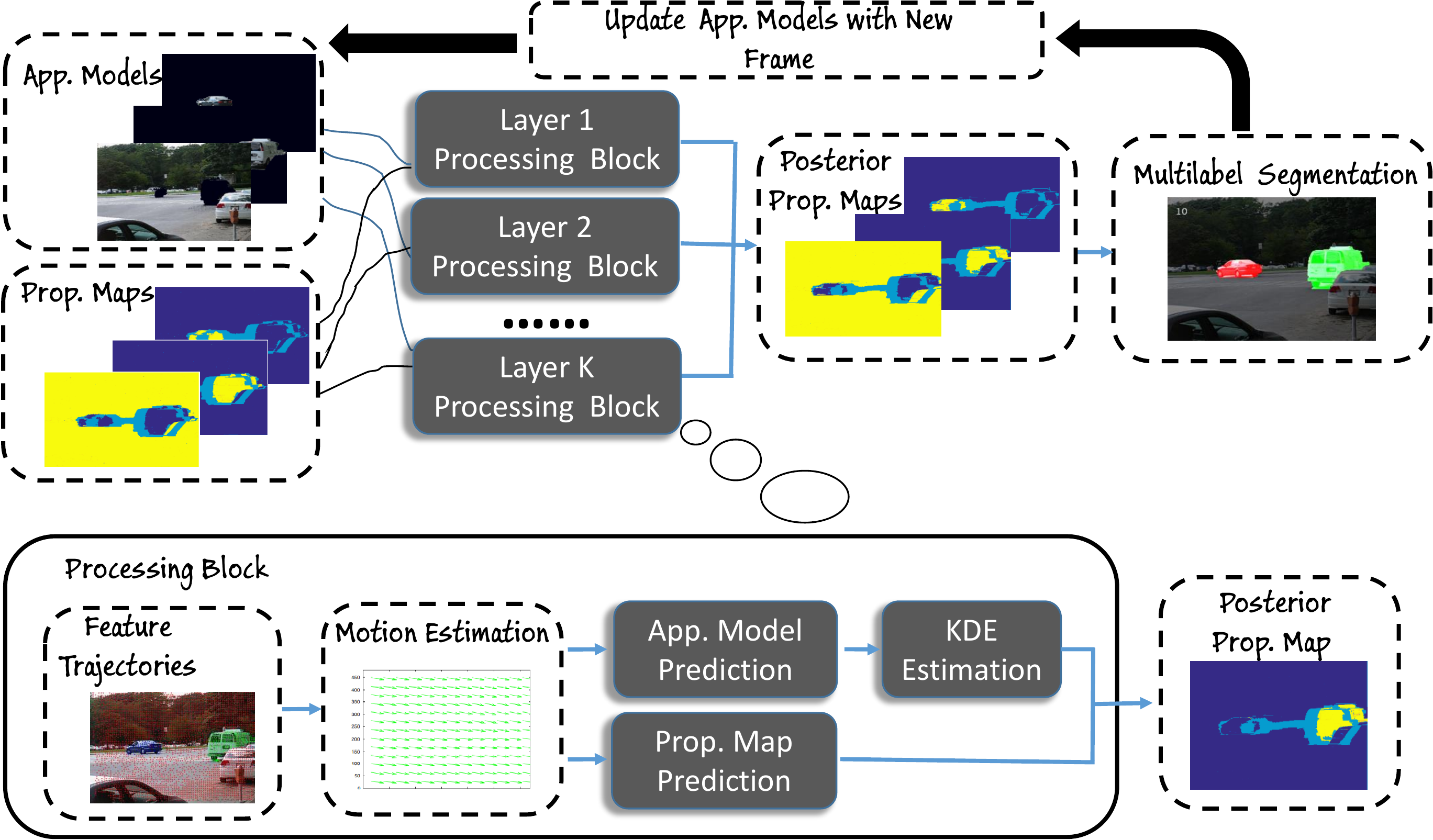}
	\caption{
		The framework of our proposed approach. 
	}
	\label{fig:framework}
\end{figure}

%When processing one frame, we have the appearance models and probability maps produced from the previews frame, which are used as the input of each corresponding layer. 

The proposed framework is demonstrated in Figure \ref{fig:framework}. 
First we employ the feature point tracking method presented in \cite{Sundaram2010} to generate feature trajectories. Then the generated trajectories are clustered into different layers based on motion inconsistency, and the labels of trajectories are continuously propagated over frames using the dynamic label propagation \cite{Zhu03LabelProp}. Each cluster of trajectories is assigned to the corresponding layer, and the number of layers is adapted according to that of foreground objects appearing in each frame.

Each layer possesses an independent ``processing block" that produces a posterior probability map. Let $k$ and $t$ denote the index of the layer and the frame respectively. 
Inside the ``processing block", we have two sources of input: the appearance model $\mathcal{A}^k_{t-1}$ and the prior probability map of labels $P(\mathcal{L}^k_{t-1})$ produced from the previous frame, and a group of corresponding sparse trajectories $\mathcal{P}^k_{t}$. The first step is the inference of the motion model in each layer using Gaussian Belief Propagation with the motion vectors of corresponding trajectories as the evidence. Then, the new appearance model $\mathcal{A}^k_t$ is obtained by shifting the previous appearance model $\mathcal{A}^k_{t-1}$ based on the estimated motion model  $\mathcal{M}^k_t$, and the new prior probability map $P_{prior}(\mathcal{L}^k_t|\mathcal{P}^k_{t})$ can be inferred from the previous one in the same way.

Given the current frame $\mathcal{I}_t$ as the new observation, the likelihood $P(\mathcal{I}_{t}| \mathcal{L}^k_{t})$ is estimated by Kernal density Estimation(KDE) \cite{Elgammal2002} with the propagated appearance model in every layer. Then the posterior probability map for each layer is inferred as 
\begin{equation}
\begin{aligned}
%P_{post}(\mathcal{L}_{t}|\mathcal{I}_t, \mathcal{P}_{t}) &\propto P(\mathcal{I}_{t}| \mathcal{L}_{t}, \mathcal{P}_{t}) P_{prior}(\mathcal{L}_t|\mathcal{P}_{t}) \\
P_{post}(\mathcal{L}^k_{t}|\mathcal{I}_t, \mathcal{P}^k_{t}) = \frac{1}{Z}P(\mathcal{I}_{t}| \mathcal{L}^k_{t}) P_{prior}(\mathcal{L}^k_t|\mathcal{P}^k_{t}), \\
%P_{post}(\mathcal{L}_{t}|\mathcal{I}_t, \mathcal{P}_{t}) & = \max_{\mathcal{L}_{t}}  P(\mathcal{I}_{t}| \mathcal{L}_{t}) P_{prior}(\mathcal{L}_t|\mathcal{P}_{t}) \\
\end{aligned}
\label{eq:infer}
\end{equation}
where $Z = \sum_k P_{post}(\mathcal{L}^k_{t}|\mathcal{I}_t, \mathcal{P}^k_{t})  $ is the partition function. With a collection of the posterior probability maps from each layer, we achieve the final pixel-wise labeling by optimizing a cut on a multilabel graph with the minimal cut energy. At the end of the whole process, appearance models are updated with the current frame and labels. In the following sections, we will describe the process steps in detail.

\section{Trajectory Labeling and Propagation}
The feature point tracking method~\cite{Sundaram2010} we employ has achived a good performance in feature trajectories generation. To cluster trajectories, several motion segmentation methods \cite{Brox2014, yan2006general} have provided good solutions, but may fail when the video does not meet the assumption of the affine camera model. To get rid of such an assumption, an online method proposed by Elqursh \etal\cite{Ali2013} considered sparse trajectory clustering as the problem of manifold separation and dynamically propagate labels over frames. Inspired by~\cite{Ali2013}, we first cluster the trajectories in the initialization frames, and continuously propagates the label of trajectories over frames using the dynamic label propagation \cite{Zhu03LabelProp}. We briefly describe the algorithm here.

\subsection{Trajectory Clustering}
Given $n$ trajectories, two distance matrices $D_M^t$ and $D_S^t$ are defined to represent the difference between trajectories in motion and spatial location. The entries $D_{ij}^M = d_M^{1:t}(T_i, T_j)$ and  $D_{ij}^S = d_S^{1:t}(T_i, T_j)$ are the distances between {\em i-th} and {\em j-th} trajoctories over frames up to $t$. For detailed defination, please refer to ~\cite{Brox2014}. The affinity matrix over $n$ trajectories is then formulated as 
\begin{equation}
\begin{aligned}
A = \exp(-(\lambda D_M^t  + (1-\lambda)D_S^t)),
\end{aligned}
\end{equation}
where $\lambda$ is the paramater to balance two distances. 

Considering each trajectory as a node and the affinity matrix as the edge weights, we cast the trajectory clustering to graph cut problem.
Starting from the initial cluster that contains all trajectories, normalized cuts~\cite{shi2000normalized} are employed to perform optimal binary cut on initial cluster and again on the generated clusters. This recursive process continues until the evaluated normalized cut cost on the cluster is above the threshold($10^{-4}$ in our work), which indicates this cluster of trajoctories belongs to the same component (i.e. objects or the background), and needs no further splitting. All trajectories are assigned with labels according to which cluster they belong to. 

\subsection{Label Propagation}
With the labels of trajectories in the intial frames, the label propagation, as a semi-supervised learning method, is adopted to infer the labels of trajectories in subsequent frames. We first construct a graph $G$ with the trajectories in the previews and current frames as the labeled and unlabeled nodes respectively. The affinity matrix $A$ involving the labeled and unlabeled trajectories are calculated and used as edge weights between corresponding nodes. 
%To predict the label of unlabeled nodes, labels are propagated on the graph. 
Let $Y_l$ and $Y_u$ denote the labeling probability matrix corresponding to labeled and unlabeled node respectively. Each row of $Y_l$ is a one-hot vector with one at the location cooresponding to the label of node and zero otherwise. To estimate the labeling probability matrix $Y_u$, we apply Markov random walks on the graph~\cite{zhu2003semi}. A transition matrix $P$ is defined as $[p_{ij} =  A_{ij}/\sum_k A_{ik} ]$, where the entry $p_{ij}$ is the transition probabilities from $ith$ node to $jth$ node. Given the partition of nodes into labeled and unlabeled nodes, the matrix P can be splitted into 4 blocks: 
\begin{equation}
\begin{aligned}
P = \begin{bmatrix}
P_{ll} &P_{lu} \\P_{ul} &P_{uu}
\end{bmatrix}.
\end{aligned}
\end{equation}

The closed form solution of the labeling probability matrix of unlabeled node is $Y_u = (I-P_{uu})^{-1}P_{ul}Y_l$. The label for the node $i$ can be obtained by
\begin{equation}
\begin{aligned}
l_i = \arg\max_j y_{ij},
\end{aligned}
\end{equation}
where $y_{ij}$ is the entry in the row corresponding to node $i$.

\begin{comment}

\begin{equation}
\begin{aligned}
d_M^{1:t}(T_a, T_b) &= \max_{i \in \{1:t\}} d^i_M(T_a, T_b),\\
d_S^{1:t}(T_a, T_b) &= \max_{i \in \{1:t\}} d^i_S(T_a, T_b)
\end{aligned}
\end{equation}
The distances for each frame are defined as: 
\begin{equation}
\begin{aligned}
d_M^i(T_a, T_b) &= \frac{(u_a^{i-\Delta:i} - u_b^{i-\Delta:i})^2 }{(\sigma^i_{Mu})^2} + \frac{(v_a^{i-\Delta:i} - v_b^{i-\Delta:i})^2 }{(\sigma^i_{Mu})^2} ,\\
d_S^i(T_a, T_b) &= \max_{i \in \{1:t\}} d^i_S(T_a, T_b)
\end{aligned}
\end{equation}

\end{comment}

It is worth noting that the label propagation algorithm can only assign the trajectories with known labels. However, when new objects move into the scene, new labels should be introduced in time. To accomplish it,  after the labels are predicted in each frame, a normalized cut cost in each cluster is evaluated. A small cost indicates a great intra-cluster variation inside the cluster. If the cost is below the threshold ($10^{-4}$ in our work), the cluster should be further splitted and a new label is assigned to the cluster with more different appearance from the previous one. When an object moves out of the scene, few trajetories are assigned with the corresponding label and the corresponding cluster is removed. In this way, the number of clusters changes adaptively according to how many moving objects appear in the scene. After clustering process is done, all trajectories in each cluster are further assigned to each layer.

\section{Motion Model Estimation}
\begin{comment}
	\begin{figure}
	\centering
	\subfloat[]{\includegraphics[width= 1.2in]{motion_traj-crop.pdf}} \quad
	\subfloat[]{\includegraphics[width= 1.2in]{motion1-crop.pdf}}\\
	\subfloat[]{\includegraphics[width= 1.2in]{motion2-crop.pdf}} \quad
	\subfloat[]{\includegraphics[width= 1.2in]{motion3-crop.pdf}}
	\setlength\belowcaptionskip{-2.5ex}
	\caption{	Visualization of labeled trajectories and motion models of three layers. The first figure is the original image covered with labeled trajectories: the trajectories are clustered into three layers: \textcolor{red}{background layer(Red)}, \textcolor{blue}{car1 layer(Blue)} and \textcolor{green}{car2 layer(Green)}. The following three figures show the estimated motion models for the three layers, where the arrow and length indicate the direction and magnitude of motion.  }
	\label{fig:motion_combo}
	\end{figure}
\end{comment}
\begin{figure}
	\centering
	\includegraphics[width= 2.5in]{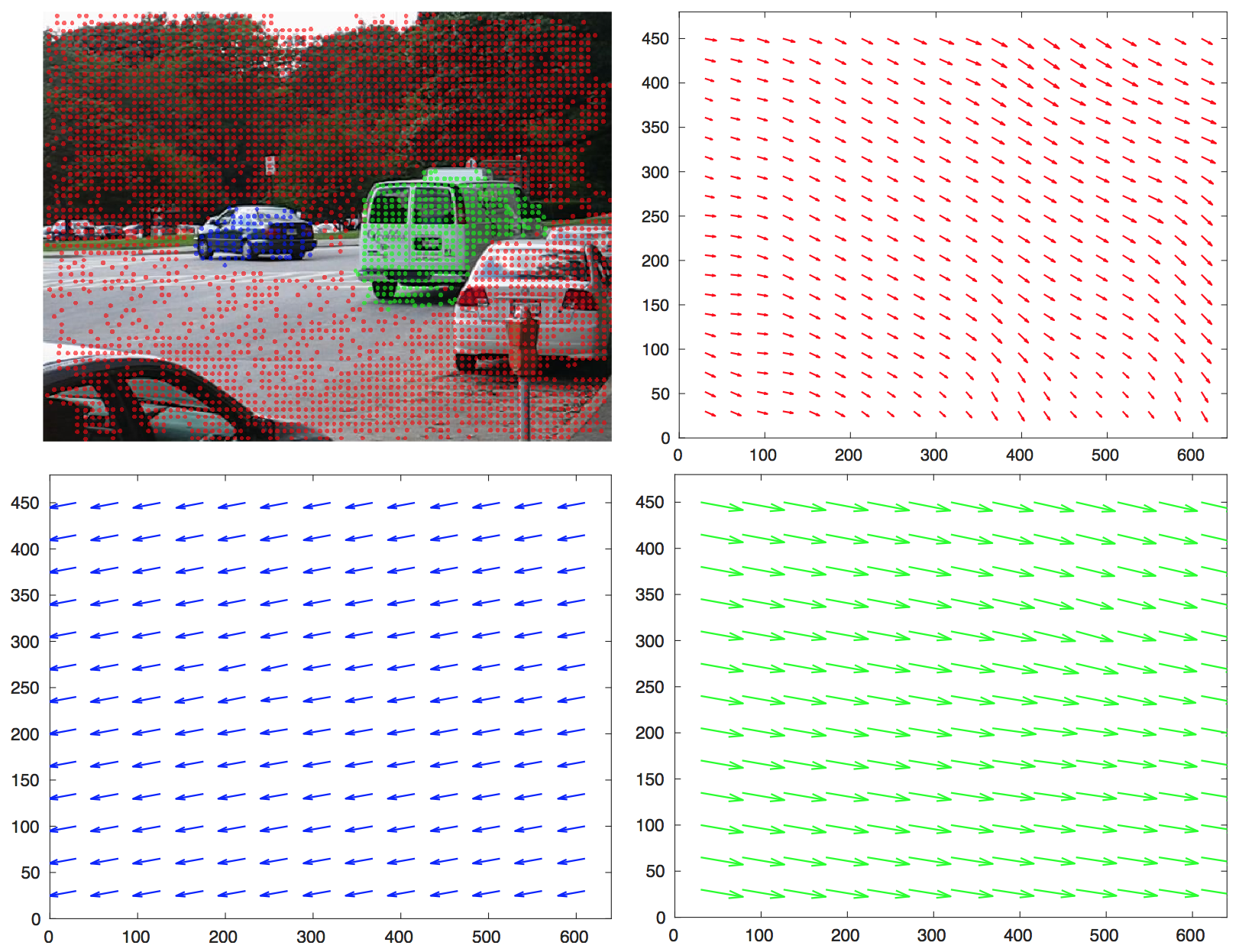}
	\caption{	Visualization of labeled trajectories and motion models of three layers. The first figure is the original image covered with labeled trajectories: the trajectories are clustered into three layers: \textcolor{red}{background layer(Red)}, \textcolor{blue}{car1 layer(Blue)} and \textcolor{green}{car2 layer(Green)}. The following three figures show the estimated motion models for the three layers, where the arrow and length indicate the direction and magnitude of motion.  }
	\label{fig:motion_combo}
\end{figure}

Although providing the motion information only at sparse pixels, trajectories of feature points are more accurate and less noisy compared with optical flow. With these accurate motion vectors of trajectories as evidence, we can estimate the motion field model for the whole frame (i.e. to estimate the motion vector of every pixel in each layer). To accomplish the pixel-wise motion estimation, we construct a pairwise MRF on a grid structure, where each vertex represents a pixel and the set of edges $\varepsilon$ represents pairwise neighborhood relationship on this structure. Two sets of potentials are involved: edge potentials $\Psi_{ij}$ measuring the similarity between neighborhood vertices, and self-potentials $\Phi_i$  measuring the likelihood of evidence. If these potentials are defined as Guassion distribution, the task is thus formulated as a Gaussian Belief Propagation(GaBP) problem \cite{GBP2008}.
Given the motion vectors of trajectories in the {\em k-th} layer, the conditional joint probability distribution of motion model can be inferred as: 
\vspace{-1mm}
\begin{equation}
P(\mathcal{M}^k_{t}|\mathcal{P}^k_{t}) \propto \prod_{(i,j)\in \varepsilon} \Psi(m_{k,t}^i,m_{k,t}^j) \prod_{i \in S_{k,t}}\Phi(m_{k,t}^i),
\end{equation}  
%\vspace{-1mm}
where $\mathcal{P}_{k,t}$ donates the set of feature points along with trajectories that are clustered to {\em k-th} layer. The edge and self potentials are defined as $\Psi(m_{k,t}^i,m_{k,t}^j) = \mathcal{N}(m_{k,t}^i|m_{k,t}^j, \Sigma_m)$ and $\Phi(m_{k,t}^i) = \mathcal{N} (m_{k,t}^i| m_{p,t}^i, \Sigma_p)$, where $m_{p,t}^i$ represents the motion vector of corresponding trajectories associated with {\em i-th} pixel. Our formulation encourages the similarity between motion vectors of neighborhood points and that between estimated motion vectors of feature points and the evidence (i.e. motion of trajectories). According to GaBP, this equation can be rewritten as
\vspace{-1mm}
\begin{equation}
P(\mathcal{M}^k_{t}|\mathcal{P}^k_{t}) \propto \exp(-\frac{1}{2}m^TAm + m^Tb),
\end{equation}  
%\vspace{-1mm}
where the inverse covariance matrix $A$ is defined to show the connection of every pair of nodes, and the shift vector $b$ is defined by the motion of trajectory. A closed form solution for marginal posterior probability is $p(m^i_{k,t}|\mathcal{P}^k_t)= \mathcal{N}(\mu^i_{k,t}, \Sigma_{k,t}^{i})$, where $\mu^i_{k,t}$ is the {\em i-th} entry of $A^{-1}b$ and  $\Sigma_{k,t}^{i}$ is the entry $\{A^{-1}\}_{ii}$. The estimated motion field is demonstrated in Figure \ref{fig:motion_combo}.

\section{Bayesian Filtering}
The inference of probability maps can be performed as the sequential Bayesian filtering on the Hidden Markov Model. 
%Besides the frame $\mathcal{I}_t$, we have motion models estimated in Section 4 as another observation. 
In this section, we first predict appearance model and prior probability for each layer given the motion models, followed by the inference of the posterior probability with new observations in the current frame. For simplicity of the expression, we remove the subscript $k$ in the rest of paper. Note that the processes in each layer are identical.

\subsection{Model Propogation}
Given the probability distribution of the estimated motion model $P(\mathcal{M}_t| \mathcal{P}_t)$, we can estimate the appearance model and the probability map in the current frame by propagating the corresponding model and map from the previous frame. Specifically, the motion vector describes exactly how a pixel shifts between the consecutive frames. Therefore, armed with the motion information of a pixel in the current frame, we can easily obtain the appearance of the pixel by propagating that of the corresponding pixel in the previous frame. However, the motion vector of each pixel has Gaussian distribution over two-dimentional space. It is impractical to marginalize the appearance over the whole Gaussian distribution. If we discard the uncertainty in the motion vector and set the variance to 0, the gaussion distribution is reduced to a Dirac delta function  $P(m^i_t| \mathcal{P}_t) = \delta(m^i_t - \mu^i_{k,t})$. Then the marginalization over the whole appearance model is reduced to the involvement of the particular pixel. The appearance model of {\em i-th} pixel  can be readily obtained by
\begin{equation}
\begin{aligned}
a^i_t =  a^{\phi(i,\mu^i_t)}_{t-1},   \\
\end{aligned}
\end{equation}
where the function $\phi(i,\mu^i_t)$ is to obtain the corresponding index in the previous frame by reversely shifting the {\em i-th}  pixel according to the associated motion vector $\mu^i_t$.  In practice, we found this approximation have little performance degradation while saving much computational cost.

Similarly, the prior probability of {\em i-th} pixel in each layer is obtained by 
\begin{equation}
\begin{aligned}
P_{prior}(l^i_t| \mathcal{P}_t) = P(l^{\phi(i, \mu_t^j)}_{t-1}).
\end{aligned}
\label{eq:labelinfer}
\end{equation}

\subsection{Probability Map Update}
%Note that since we use the KDE method to model the appearance, the appearance model of pixel i $a_i$ is involved with a pool of pixels accumulated from the previews frames. The details will be shown later. 
Once the appearance model and probability map are propagated from the previous frame, the posterior probability map of each layer can be inferred with the current Frame $\mathcal{I}_t$ as the new observation. With the assumption of independence between pixels, the posterior probability of {\em i-th} pixel is computed by Eq. (\ref{eq: postelement}). 
\begin{equation}
P_{post}(l^i_t| I^i_t, \mathcal{P}_t) \propto P(I^i_t| l^i_t) P_{prior}(l^i_t|\mathcal{P}_t),
\label{eq: postelement}
\end{equation}

The likelihood of the observation in {\em i-th} pixel $P(I^i_t|l^i_t)$ is essentially describing how well the appearance $I^i_t$ fits the appearance model in each layer . Kernel Density Estimation(KDE), as a nonparametric method, can effectively model the appearance for background or foreground without the restriction of hyper-parameter estimation.  For its simplicity and effectiveness, we employ KDE technique for appearance modeling. The appearance model of  {\em i-th} pixel $a^i_t$ is involved with a pool of color features $\{I_f\}^i_N$ accumulated from the previews frames. The likelihood of the observation $P(I^i_t|l^i_t)$ estimated as:

\begin{equation}
P(I^i_t|l^i_t) = \frac{1}{N} \sum_{f=1}^{N}  K_G(I_t^i- I_{f}^i),
\end{equation}
where $K_G(\cdot)$ is the Gaussian kernel function, $I_f^i$ is the color feature in $f$ frame stored for KDE modeling, and $N=20$ is the number of stored previous frames. The posterior probability map produced in each layer is normalized by partition function, and then used as the knowledge for multilabel segmentation.

\begin{figure}
	\centering
	\includegraphics[width= 2.5in]{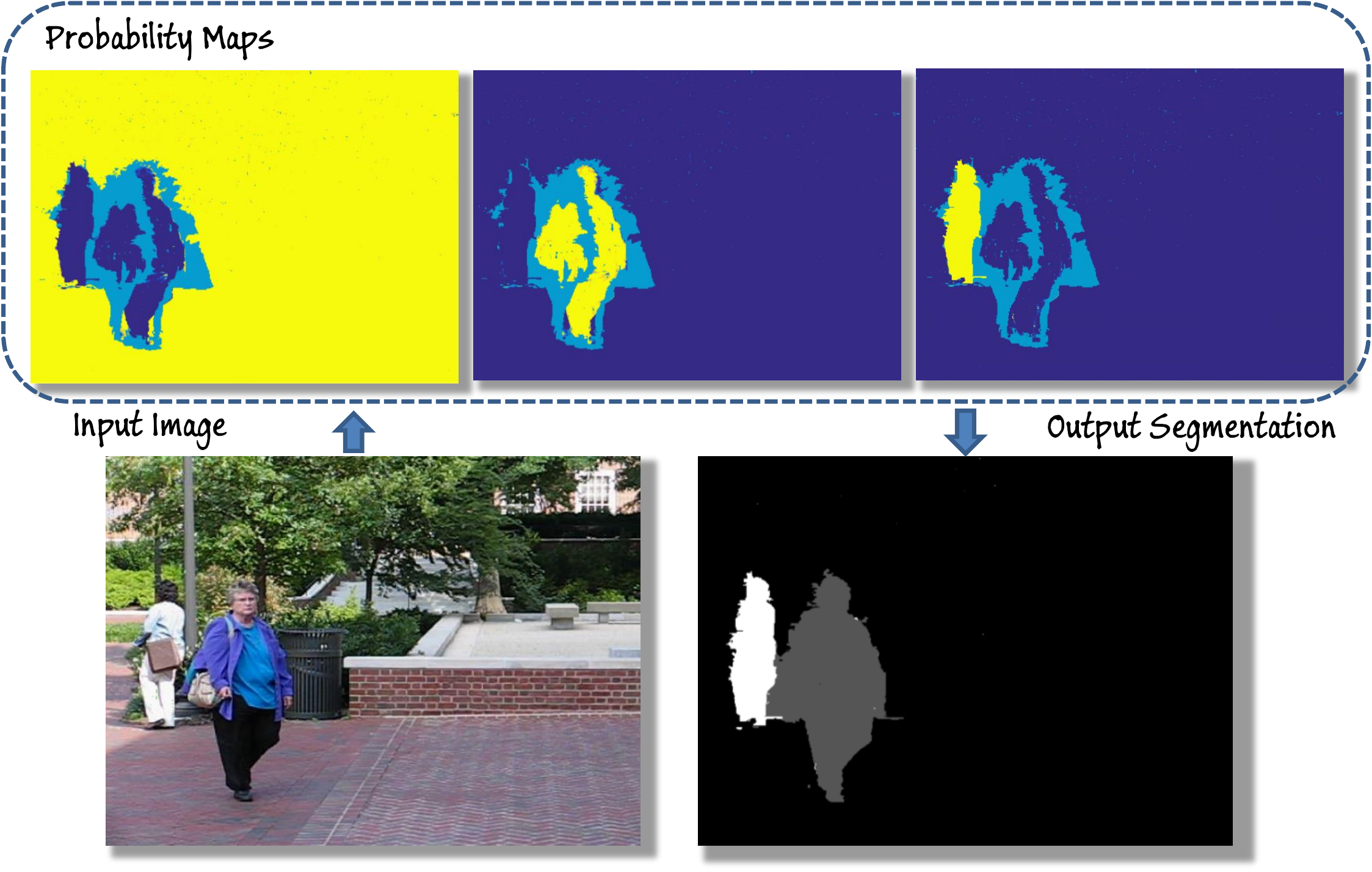}
	\setlength\belowcaptionskip{-2.5ex}
	\caption{Inside the dotted box, three probability maps are demonstrated: the background layer, foreground1 layer(woman on the right), foreground2 layer(woman on the left). Note that the lighter color represents the high confidence. Two images in the bottom show the input image and the result of Multilabel Segmentation. }
	\label{fig:applresult}
\end{figure}

\section{Multilabel Segmentation}
With the collection of normalized probability maps of all layers at hand, our final task for foreground objects segmentation is to perform pixel-wise labeling for the whole frame. This segmentation problem can be converted into an energy minimization problem on the pairwise MRF which can be polynomially solvable via Graph Cuts \cite{Graphcuts}. Due to ill-posed nature of the segmentation problem, regularizations are always required. For our problem, we designed two regularizers: a smoothing cost and a label cost in preference of smoothness of labeling and fewer unique labels assigned, respectively. The global energy function is formulated as: 
\begingroup\makeatletter\def\f@size{9}\check@mathfonts
\def\maketag@@@#1{\hbox{\m@th\large\normalfont#1}}
\begin{equation}
\begin{aligned}
E(\mathcal{L}) = & \sum_{k\in {S(\mathcal{L})}}\sum_{i:l_i =k} E(I_i) + \lambda_1 \sum_{(i,j) \in \mathcal{N}} q_{ij}E(I_i,I_j) \\
&+ \lambda_2 \sum_{k\in {S(\mathcal{L})}}h_k\delta_k(\mathcal{L})
\end{aligned}
\end{equation}
\endgroup
where $S(\mathcal{L})$ denotes the set of unique labels of $\mathcal{L}$. The three terms in the right-side of equation are data cost, smoothness cost and label cost, respectively. The data cost $E(I_i)$ is defined as the negative log probability of $ith$ pixel belonging to a certain layer $-\log P_{post}(l^i_t| I^i_t)$. The smoothness cost is defined as the similarity of two neighboring pixels $E(I_i,I_j) = -\log \mathcal{N}(I_i | I_j, \Sigma_a)$, and $q_{ij}$, as a sign function, equals to $+1$ if $l_i$, $l_j$ have the same label; otherwise $-1$. Moreover, in the term of label cost, $h_k$ is the non-negative label cost of the label $k$. And $\delta_k(f)$ is the corresponding indicator function with the definition $\delta_k(\mathcal{L}) = 1$ if $\exists i:l_i =k;$ and $0$ otherwise. $\lambda_1, \lambda_2$ are non-negative parameters to balance data cost against such two regularizations (both are 1 in our work). The energy minimization function can be solved efficiently using the method presented in \cite{GraphcutsLabelCosts}. 

It is worth noting that when a new foreground object appears in the scene or initial frames are processed, the feature samples to build the appearance model usually are not sufficient for Kernel Density Estimation. In these cases that KDE is invalid, the probability maps is no longer avaiable as the data cost for the multilabel segmentation. An alternative way to define the term of data cost in energy function is based on the labeled trajectories. 
$E(I_i) = c$ if $l_i = l_p $ and $-c$ otherwise, where $c$ is a negative constant, and $l_p$ is the label of trajectory associated with {\em i-th} pixel. This defination simply ultilizes the motion informance rather than the appearance model.

Finally, according to the labels of pixels, the appearance models $\mathcal{A}_{t}$  are updated by adding the color feature $I_t^i$ in the current frame to the corresponding appearance model.

\section{Experiments}
We evaluate our algorithm qualitatively and quantitatively based on two criteria: one is the normal two-label background subtraction and one is multilabel background segmentation. The former is evaluated using F-score, which is the common measurement. For the latter, since, to our best knowledge, no one has done such work before, we carefully design a reasonable measurement. The result shows our method outperforms the state-of-the-art approaches in both settings.

%Actually, there is no benchmark dataset exactly designed for performance evaluation of background subtraction methods for moving cameras. 
\vspace{1.2mm}
\noindent 
\textbf{Dataset:} Experiments were run on two public datasets. A set of sequences (cars1-8, people1-2) in the Hopkins dataset~\cite{hopkin2007} is commonly used for quantitative evaluation on this topic, some of which contain multiple foreground objects. To quantitatively evaluate the performance, we produced the groundtruth mask manually for all frames, including the discrimination of foreground objects. Another one is Complex Background Dataset(CBD)~\cite{narayana2013}, where the complex background and camera rotation introduce a great challenge.   

\setlength{\textfloatsep}{10pt plus 1.0pt minus 1.0pt}
\setlength{\tabcolsep}{3pt}
\begin{table}[h]
	%\begin{center}
	\centering 
	\begin{tabular}{c|cccccc}
		\hline
		%	& \multicolumn{1}{c}{ours MLBS}  & \multicolumn{1}{c}{ Lim \textit{et al.} \cite{Lim2014}} & \multicolumn{1}{c}{Elqursh \textit{et al.} %\cite{Ali2012}}  & \multicolumn{1}{c}{kwak \textit{et al.} \cite{han2011}}  &\multicolumn{1}{c}{Sheikh \textit{et al.} \cite{Sheikh2009}}\\
		&&&&&\\[-0.8em]
		& \multicolumn{1}{c}{MLBS}  & \multicolumn{1}{c}{GBSSP} & \multicolumn{1}{c}{FOF} & \multicolumn{1}{c}{OMCBS}  & \multicolumn{1}{c}{GBS}  &\multicolumn{1}{c}{BSFMC}\\
		\hline
		%\cline{2-4} \cline{5-7} \cline{8-10} \cline{11-13}
		% & F           & F   & F & F\\ \hline
		\hline
		&&&&&\\[-0.8em] 
		cars1     &  \textbf{92.04} 	&  87.14 &50.84  &91.77 & 80.30 &73.13\\	%\hline
		cars2      &  \textbf{90.16}	&  82.17 &56.60 &69.13	& 68.45 &55.68\\	%\hline
		cars3     &  \textbf{93.16}     &  72.94 &73.57  &41.27 & 79.22 &60.91\\	%\hline
		cars4     &  \textbf{91.55}	    &  88.24 &47.96 &73.65	& 66.63 &54.81 \\	%\hline
		cars5    	& \textbf{86.62}    & 81.66  &70.94 &60.44	& 74.56 &51.97\\	%\hline
		cars6     & \textbf{92.23}	   & 81.44  &84.34 &90.31	&73.34 &37.56\\	%\hline
		cars7      & \textbf{91.17}  	& 90.86 &42.92 & 89.87	& 69.10 &40.35\\	%\hline
		cars8       & 85.93		& \textbf{86.85} &87.61 & 83.84& 80.29 &62.42\\	%\hline
		people1     &  \textbf{81.38}	& 81.21 &69.59 &64.04	& 80.19 &34.25\\	%\hline
		people2      & \textbf{94.34}	&84.74 &88.40 &89.32	& 81.40 &64.92\\	\hline \hline
		&&&&&\\[-0.8em]
		drive     &  \textbf{65.95} 	&  53.55 &61.80 &13.68 & 5.18 &2.02\\	%\hline
		forest    &  72.20  	&\textbf{91.44} &31.44 &42.99 & 23.19 &16.76\\	%\hline
		parking   &  \textbf{83.66}   &  68.97 &73.19 &11.47 & 11.02 &13.05\\	%\hline
		store     &  \textbf{86.28}	    & 83.44 &70.74 &10.18 & 21.42 &8.83 \\	%\hline
		traffic   & 48.19	&  31.31     &\textbf{71.24} &41.49 & 24.14 &27.49\\	\hline \hline
		
		&&&&&\\[-0.8em]
		Overall    & \textbf{83.66}	  &77.73 &65.45 & 58.23	& 55.90 & 40.28\\	\hline
		%Overall    & \textbf{89.86}	  &83.72 &67.27 & 75.36	& 75.35 & 53.60\\	\hline
		
	\end{tabular}
	\caption{
		Two-label background subtraction performance comparison with the F-score(\%) on \textbf{Hopkins} and \textbf{CBD} Dataset. Best performance scores are highlighted in bold.
	}
	\label{tb:quantitative1}
\end{table}

\begin{comment}
\setlength{\tabcolsep}{3pt}
\begin{table}[h]
	%\begin{center}
	%\vspace{-4mm}
	\centering 
	\newcolumntype{.}{D{.}{.}{-1}}
	\begin{tabular}{c|c c c c . .}
		\hline
		%	& \multicolumn{1}{c}{ours MLBS}  & \multicolumn{1}{c}{ Lim \textit{et al.} \cite{Lim2014}} & \multicolumn{1}{c}{Elqursh \textit{et al.} %\cite{Ali2012}}  & \multicolumn{1}{c}{kwak \textit{et al.} \cite{han2011}}  &\multicolumn{1}{c}{Sheikh \textit{et al.} \cite{Sheikh2009}}\\
		&&&&&\\[-0.8em]
		& \multicolumn{1}{c}{MLBS}  & \multicolumn{1}{c}{GBSSP} & \multicolumn{1}{c}{FOF} & \multicolumn{1}{c}{OMCBS}  & \multicolumn{1}{c}{GBS}  &\multicolumn{1}{c}{BSFMC}\\
		\hline
		%\cline{2-4} \cline{5-7} \cline{8-10} \cline{11-13}
		% & F           & F   & F & F\\ \hline
		\hline
		&&&&&\\[-0.8em]
		drive     &  \textbf{65.95} 	&  53.55 &61.80 &13.68 & 5.18 &2.02\\	%\hline
		forest    &  72.20  	&\textbf{91.44} &31.44 &42.99 & 23.19 &16.76\\	%\hline
		parking   &  \textbf{83.66}   &  68.97 &73.19 &11.47 & 11.02 &13.05\\	%\hline
		store     &  \textbf{86.28}	    & 83.44 &70.74 &10.18 & 21.42 &8.83 \\	%\hline
		traffic   & 48.19	&  31.31     &\textbf{71.24} &41.49 & 24.14 &27.49\\	\hline\hline
		&&&&&\\[-0.8em]
		Overall    & \textbf{71.25}     &  65.74 &61.82 & 23.96  &16.99 & 13.63\\	\hline
		
	\end{tabular}
	\caption{
		Performance comparison with the F-score(\%) on \textbf{CBD} Dataset. 
	}
	\label{tb:quantitative2}
\end{table}
\end{comment}

\setlength{\tabcolsep}{1.0pt}
\begin{figure*}
	\centering
	\includegraphics[scale = 0.7]{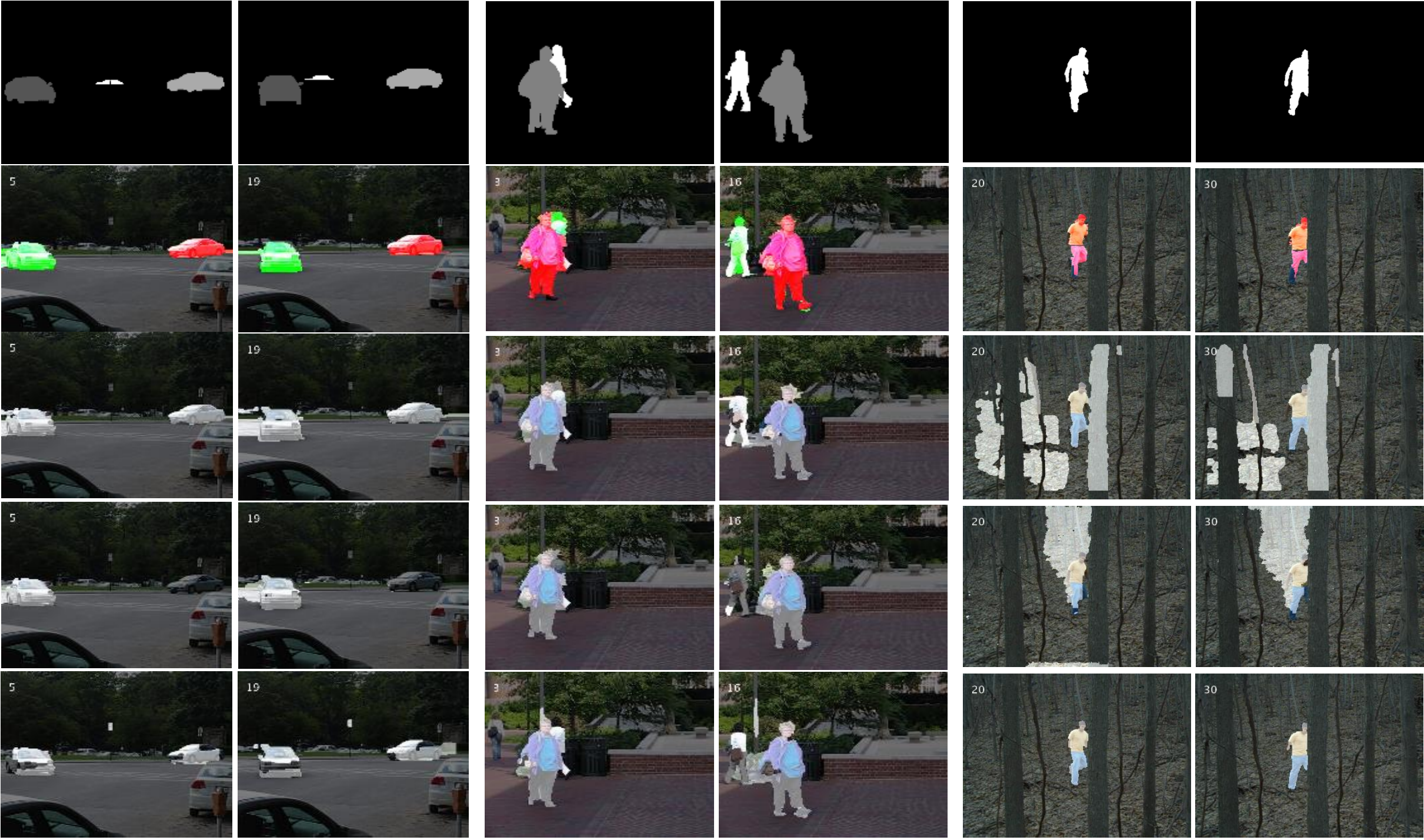}
	\setlength\belowcaptionskip{-2.5ex}
	\caption{
		Comparison of our MLBS, and three state-of-the-art algorithms on the sequences cars5, people2 and forest, separated in three blocks. Row1: Groundtruth; Row2: our MLBS; Row3: GBS~\cite{han2011}; Row4: OMCBS~\cite{Ali2012}; Row5: GBSSP~\cite{Lim2014}. For clear demonstration, the background is darkened and the foreground objects are marked in different colors. (Better seen in color)
	}
	\label{fig:segcompare}
\end{figure*}
\subsection{Two-label Background Subtraction}

The performance of our framework, represented as MLBS (Multi-Layer Background Subtraction), is compared to five state-of-the-art algorithms:
%Lim \textit{et al.} \cite{Lim2014},  Kwak \textit{et al.} \cite{han2011} and Sheikh \textit{et al.} \cite{Sheikh2009}. 
GBSSP \cite{Lim2014}, FOF \cite{narayana2013}, OMCBS \cite{Ali2012}, GBS \cite{han2011}, and BSFMC \cite{Sheikh2009}. GBS requires the initialization labels of each frame as additional inputs,  and GBSSP needs the groundtruth of the first frame as the initialization. For fair comparisons, we provide GBS with the labeling results of BSFMC, and offer GBSSP the labeling result of the first frame generated by our method. Note that instead of the requirement of these additional informations, our MLBS method completes self-initialization automatically.
 %(Details about initialization are shown in the Supplementary Material). 
 Moreover, we use the parameters provided by authors in external methods and fixed them for each algorithm in all experiments.

% and Zamalieva \textit{et al.} \cite{Zamalieva2014}

The quantitative comparisons on two dataset are shown in Table \ref{tb:quantitative1}. It can be seen that the proposed method outperforms other methods in the literature on most test sequences. Especially in the case where the objects are occluded and then separated, such as the cars3 and people2, the multilayer strategy boosts the performance with a great jump on F-score. This is because our method can accurately estimate separate motion models for foreground objects, instead of only one ambiguous model for the whole foreground, and separate appearance models comprehensively provide the evidence of probabilistic inference. The overall F-score of our method is higher than the best score of the state-of-the-arts by a noticeable margin (83.66\% vs 77.73\%).
% the best scores among the previous state-of-the-art methods.

%\vspace{-1mm}

To evaluate the ability of the proposed method handling multiple objects, we categorize the suquences into three scenarios according to the number of moving objects in the scene, and report the average F-score on each scenario in Table~\ref{tb:3class}. In the single object scenario, our method outperform the second best method by a small margin (84.05\% vs 80.70\%). But as the number of objects increases, the margin grows (3.4\% vs 8.0\% vs 8.7\%). It clearly demonstrates the outstanding capability of our method in complex scenes.

\setlength{\tabcolsep}{3pt}
\begin{table}[h]
	%\begin{center}
	\centering 
	\begin{tabular}{c|cccccc}
		\hline
		%	& \multicolumn{1}{c}{ours MLBS}  & \multicolumn{1}{c}{ Lim \textit{et al.} \cite{Lim2014}} & \multicolumn{1}{c}{Elqursh \textit{et al.} %\cite{Ali2012}}  & \multicolumn{1}{c}{kwak \textit{et al.} \cite{han2011}}  &\multicolumn{1}{c}{Sheikh \textit{et al.} \cite{Sheikh2009}}\\
		&&&&&\\[-0.8em]
		& \multicolumn{1}{c}{MLBS}  & \multicolumn{1}{c}{GBSSP} & \multicolumn{1}{c}{FOF} & \multicolumn{1}{c}{OMCBS}  & \multicolumn{1}{c}{GBS}  &\multicolumn{1}{c}{BSFMC}\\
		\hline
		%\cline{2-4} \cline{5-7} \cline{8-10} \cline{11-13}
		% & F           & F   & F & F\\ \hline
		\hline
		&&&&&\\[-0.8em] 
		$1$      &  \textbf{84.05} 	&  80.70 &59.20  &54.22 & 47.82 &31.20\\	%\hline
		$2$       &  \textbf{91.14}	& 81.51 &83.19 &71.48	& 80.30 & 62.75\\	%\hline
		$\geq3$    &  \textbf{74.99}     &  65.05 &66.26  &57.02 & 55.72 &45.05\\	%\hline
		\hline

		%Overall    & \textbf{89.86}	  &83.72 &67.27 & 75.36	& 75.35 & 53.60\\	\hline
		
	\end{tabular}
	\caption{
		Two-label background subtraction performance comparison on the videos with different numbers of moving objects. Best performance scores are highlighted in bold.
	}
	\label{tb:3class}
\end{table}

Qualitative comparisons with three algorithms are illustrated in Figure \ref{fig:segcompare}. We pick two representative sequences (cars5 and people2), where at least two foreground objects appear in the scene, from Hopkins Dataset, and one sequence (forest) from CBD.  
It is notable that our MLBS algorithm separates moving objects and background accurately when the background scene is complex and the motions of objects are not simply in the same direction. For instance, in the sequence of people2, two women, who are walking in different directions, occluded and then separated, are separated precisely while other methods wrongly label the new appearing background region (e.g. the black rubbish bin) as foreground.

%\vspace{-1mm}
\subsection{Multilabel Background Subtraction}

%Besides the performance evaluation of binary separation of background and foreground, 
We also evaluate the capability to separate different foreground objects. Since no one, to our best knowledge, has done such work before, we have designed a baseline to compare the performance. Details about baseline are shown in the Supplementary Material. Besides, \cite{Brox2014} have proposed an offline approach (SLT) to tackle video segmentation problem based on long-term video analysis, and achieved a cutting edge performance. Since this method has no discrimination of foreground and background, we modify it by manually assigning the background label to the best fit segmented region, and make a comparison with it.

For quantitative comparison, we design a measurement by modifying the metric used in \cite{Brox2014}. 
Let $g_i$ denote the foreground region in the groundtruth, $c_i$  the corresponding regions in the mask generated by algorithms, and $|.|$ the number of pixels inside the region. For each foreground region in the groundtruth, precision, recall and F-score are defined as:
\vspace{-1mm}
\begin{equation}
P_i = \frac{|c_i \cap g_i|}{|c_i|}\quad  R_i = \frac{|c_i \cap g_i|}{|g_i|} \quad F_i = \frac{2P_iR_i}{P_i + R_i}
\end{equation}
\vspace{-1mm}

Overall metric is obtained by averaging the measures of single regions. And the best one-to-one assignment of generated regions to groundtruth regions is found by the Hungarian method \cite{kuhn1955hungarian}. In the case where there exist generated regions without the assignment of groundtruth regions, we set the precision and recall of such regions to 1 and 0 respectively. It's worth mentioning that our revised metric is calculated over only foreground regions to keep consistency with the binary background subtraction, as the measurement values are the same as that of the binary one when there is only one foreground object in the scene. 
%the previous metric \cite{Brox2014} takes into consider all segmented regions, and 

\setlength{\tabcolsep}{4.0pt}
\begin{table}
	\vspace{-2mm}
	\centering
	\begin{tabular}{>{\centering}m{1.2cm}c|c c c|}
		\cline{3-5}	
		\multicolumn{2}{ c| }{} &\multicolumn{3}{ c| }{} \\[-0.8em] 	
		& & Precision & Recall & F-Score \\ \hline \hline %\cline{1-5} \cline{1-5}
		\multicolumn{1}{ |c  }{\multirow{3}{*}{ } } &
		\multicolumn{1}{ |c| }{} & & &\\[-0.8em]   %%% this line is just to adjust the height of row
		\multicolumn{1}{ |c  }{\multirow{3}{*}{First 10 Frames } } &
		\multicolumn{1}{ |c| }{ours}  &89.79	&\textbf{83.00} & \textbf{85.06}     \\ %\cline{2-5}
		\multicolumn{1}{ |c  }{}                        &
		\multicolumn{1}{ |c| }{SLT} 	&\textbf{90.70} &78.94 &83.41   \\% \cline{2-5}
		\multicolumn{1}{ |c  }{}                        &
		\multicolumn{1}{ |c| }{Baseline} 	& 61.23 & 62.32 & 61.35    \\ \cline{1-5}
		
		\multicolumn{1}{ |c  }{\multirow{3}{*}{ } } &
		\multicolumn{1}{ |c| }{} & & &\\[-0.8em] 
		\multicolumn{1}{ |c  }{\multirow{3}{*}{	All Frames} } &
		\multicolumn{1}{ |c| }{ours} &88.60	&\textbf{85.05} & \textbf{86.16} \\ %\cline{2-5}
		\multicolumn{1}{ |c  }{}                        &
		\multicolumn{1}{ |c| }{SLT} 	&\textbf{89.51} &78.24 &82.00  \\%\cline{2-5}
		\multicolumn{1}{ |c  }{}                        &
		\multicolumn{1}{ |c| }{Baseline} 	& 63.63 & 66.22 & 64.35  \\ \cline{1-5}
	\end{tabular}
	\caption{
		Performance comparison of Multi-foreground segmentation on \textbf{Hopkins} Dataset
	}
	\label{table:Multi}
\end{table}

The performance evaluation is shown in Table \ref{table:Multi}. We have compared the performance of the approaches evaluated on both all frames and the first ten frames. It can be seen that the proposed method outperforms SLT on both sets and has a great jump from the baseline. It's notable that the F-score of our method on all frames is higher than that on first ten frames due to the nature of online methods. Unlike the offline methods that hold the knowledge of the whole sequence at the beginning of the sequence process, our online approach has little prior knowledge and requires the self-initialization step during the first several frames, which surely leads to lower performance due to the insufficiency of prior knowledge. But it affects little as the sequence becomes longer.

\begin{figure}[t]
	\centering
	\includegraphics[width=0.48\textwidth]{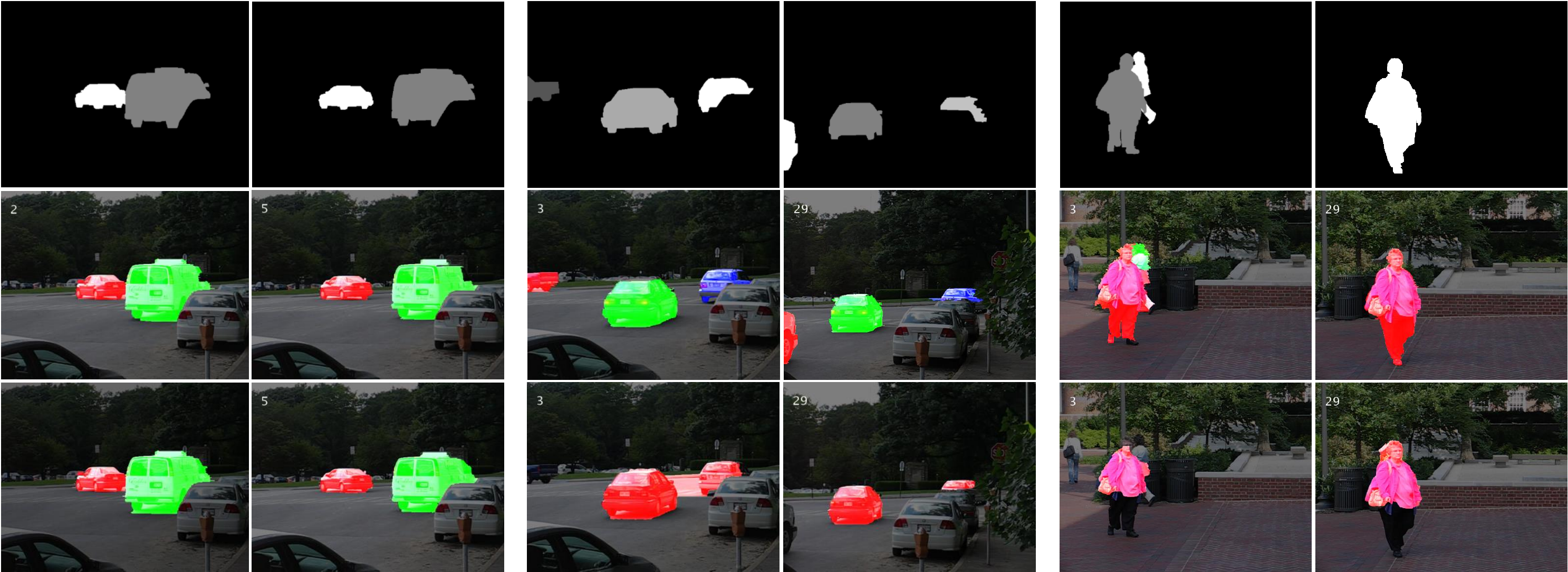}
	%\vspace{-10cm}
	\caption{
		The Multilabel segmentation performance comparison with SLT on the sequence cars3, cars2 and people2, separated in three blocks. Row1: Groundtruth; Row2: MLBS; and Row3: SLT~\cite{Brox2014}. (Better seen in color)
	}
	\label{fig:broxcompare}
\end{figure}

Qualitative evaluations of our method and SLT are demonstrated in Figure \ref{fig:broxcompare}. Our proposed method can separate the objects more accurately while SLT may falsely recognize two objects as only one when objects are very near and moving in the similar direction (see Block$2$). Furthermore, our method could detect new objects immediately when they enter the scene (see Block$2$). With the ability of accurate and robust foreground objects detection and segmentation, our method produces a proper number of foreground object regions, which is reflected by the higher recall in Table \ref{table:Multi}. Additionally, equipped with appearance models for each layer, our MLBS method is capable of dealing with the articulated motion (see Block$3$) to a certain extent.

%\begin{comment}
\setlength{\tabcolsep}{5pt}
\begin{table}[h]
	%\begin{center}
	\centering 
	\begin{tabular}{|c|c|c|c|c|c|}
		\hline 	
	tclb &mtest & mdprop& grapcut& upmdl & Total	\\[+0.2em]
	
		\hline
	%	&&&&&\\[-0.8em]
		 1109	&  3675 &144  &1369 &559 & 6856\\	%\hline
		%\hline
		\hline
		
	\end{tabular}
	\caption{
		Average computation time (msec) over different stages on a single frame from cars1. The stages are: trajectory clustering and label propagation (\textbf{tclb}), motion estimation(\textbf{mtest}), model propogation (\textbf{mdprop}), graphCut (\textbf{grapcut}) and update model(\textbf{upmdl}). 
	}
	\label{tb:time}
\end{table}

Table~\ref{tb:time} shows the average computational time per frame. Our un-optimized Matlab implementation takes around 7 seconds per frame with an Intel Xeon-E5 CPU and 16GB menory. The computational time is dominated by motion estimation and graphcut. Motion estimation is done mainly by matrix inversion and multiplication. Since such operations can be readily parallelized on the GPU, we believe the real-time performance can be achieved by the optimation with the GPU and faster multi-thread implementation. 
%\end{comment}

\section{Conclusion}
We propose a novel online multilayer-based framework for background subtraction with moving camera. In our framework, every foreground object and the background are assigned to an independent processing layer. A processing block is carefully designed to perform the posterior inference using Bayesian Filtering Framework, and Multi-label Graph-cut is employed to produce the pixel-wise segmentation for every video frame based on the normalized probability maps. Experiments show that our method performs favorably against other state-of-the-art methods, with outstanding ability to segment multiple foreground objects.

{\small
\bibliographystyle{ieee}
\bibliography{egbib}

\begin{thebibliography}{10}\itemsep=-1pt

\bibitem{GBP2008}
D.~Bickson.
\newblock Gaussian belief propagation: Theory and aplication.
\newblock {\em CoRR}, abs/0811.2518, 2008.

\bibitem{bideau2016s}
P.~Bideau and E.~Learned-Miller.
\newblock It's moving! a probabilistic model for causal motion segmentation in
  moving camera videos.
\newblock {\em arXiv preprint arXiv:1604.00136}, 2016.

\bibitem{chan2008modeling}
A.~B. Chan and N.~Vasconcelos.
\newblock Modeling, clustering, and segmenting video with mixtures of dynamic
  textures.
\newblock {\em IEEE transactions on pattern analysis and machine intelligence},
  30(5):909--926, 2008.

\bibitem{chan2009layered}
A.~B. Chan and N.~Vasconcelos.
\newblock Layered dynamic textures.
\newblock {\em IEEE Transactions on Pattern Analysis and Machine Intelligence},
  31(10):1862--1879, 2009.

\bibitem{Cui2012}
X.~Cui, J.~Huang, S.~Zhang, and D.~N. Metaxas.
\newblock Background subtraction using low rank and group sparsity constraints.
\newblock In {\em Proceedings of the 12th European Conference on Computer
  Vision - Volume Part I}, pages 612--625, 2012.

\bibitem{GraphcutsLabelCosts}
A.~Delong, A.~Osokin, H.~N. Isack, and Y.~Boykov.
\newblock Fast approximate energy minimization with label costs.
\newblock {\em International journal of computer vision}, 96(1):1--27, 2012.

\bibitem{Elgammal2014}
A.~Elgammal.
\newblock {\em Background Subtraction: Theory and Practice}.
\newblock Morgan \& Claypool Publishers, 2014.

\bibitem{Elgammal2002}
A.~Elgammal, R.~Duraiswami, D.~Harwood, and L.~S. Davis.
\newblock Background and foreground modeling using nonparametric kernel density
  estimation for visual surveillance.
\newblock {\em Proceedings of the IEEE}, 90(7):1151--1163, Jul 2002.

\bibitem{Ali2012}
A.~Elqursh and A.~Elgammal.
\newblock Online moving camera background subtraction.
\newblock In {\em European Conference on Computer Vision}, pages 228--241.
  Springer, 2012.

\bibitem{Ali2013}
A.~Elqursh and A.~Elgammal.
\newblock Online motion segmentation using dynamic label propagation.
\newblock In {\em Proceedings of the IEEE International Conference on Computer
  Vision}, pages 2008--2015, 2013.

\bibitem{Fischler1981}
M.~A. Fischler and R.~C. Bolles.
\newblock Random sample consensus: A paradigm for model fitting with
  applications to image analysis and automated cartography.
\newblock {\em Commun. ACM}, 24(6):381--395, June 1981.

\bibitem{Han2012}
B.~Han and L.~S. Davis.
\newblock Density-based multifeature background subtraction with support vector
  machine.
\newblock {\em IEEE Transactions on Pattern Analysis and Machine Intelligence},
  34(5):1017--1023, May 2012.

\bibitem{Hartley2004}
R.~I. Hartley and A.~Zisserman.
\newblock {\em Multiple View Geometry in Computer Vision}.
\newblock Cambridge University Press, ISBN: 0521540518, second edition, 2004.

\bibitem{hayman2003}
E.~Hayman and J.-O. Eklundh.
\newblock Statistical background subtraction for a mobile observer.
\newblock In {\em Computer Vision, 2003. Proceedings. Ninth IEEE International
  Conference on}, pages 67--74. IEEE, 2003.

\bibitem{jin2008}
Y.~Jin, L.~Tao, H.~Di, N.~I. Rao, and G.~Xu.
\newblock Background modeling from a free-moving camera by multi-layer
  homography algorithm.
\newblock In {\em 2008 15th IEEE International Conference on Image Processing},
  pages 1572--1575. IEEE, 2008.

\bibitem{kim2005background}
K.~Kim, D.~Harwood, and L.~S. Davis.
\newblock Background updating for visual surveillance.
\newblock In {\em International Symposium on Visual Computing}, pages 337--346.
  Springer, 2005.

\bibitem{Graphcuts}
V.~Kolmogorov and R.~Zabin.
\newblock What energy functions can be minimized via graph cuts?
\newblock {\em IEEE Transactions on Pattern Analysis and Machine Intelligence},
  26(2):147--159, Feb 2004.

\bibitem{kuhn1955hungarian}
H.~W. Kuhn.
\newblock The hungarian method for the assignment problem.
\newblock {\em Naval research logistics quarterly}, 2(1-2):83--97, 1955.

\bibitem{kumar2008learning}
M.~P. Kumar, P.~H. Torr, and A.~Zisserman.
\newblock Learning layered motion segmentations of video.
\newblock {\em International Journal of Computer Vision}, 76(3):301--319, 2008.

\bibitem{han2011}
S.~Kwak, T.~Lim, W.~Nam, B.~Han, and J.~H. Han.
\newblock Generalized background subtraction based on hybrid inference by
  belief propagation and bayesian filtering.
\newblock In {\em 2011 International Conference on Computer Vision}, pages
  2174--2181, 2011.

\bibitem{DarShyangLee2005}
D.-S. Lee.
\newblock Effective gaussian mixture learning for video background subtraction.
\newblock {\em IEEE Transactions on Pattern Analysis and Machine Intelligence},
  27(5):827--832, May 2005.

\bibitem{Lim2014}
J.~Lim and B.~Han.
\newblock Generalized background subtraction using superpixels with label
  integrated motion estimation.
\newblock In {\em European Conference on Computer Vision}, pages 173--187.
  Springer, 2014.

\bibitem{Mittal2000}
A.~Mittal and D.~Huttenlocher.
\newblock Scene modeling for wide area surveillance and image synthesis.
\newblock In {\em Computer Vision and Pattern Recognition, 2000. Proceedings.
  IEEE Conference on}, volume~2, pages 160--167 vol.2, 2000.

\bibitem{mumtaz2014joint}
A.~Mumtaz, W.~Zhang, and A.~B. Chan.
\newblock Joint motion segmentation and background estimation in dynamic
  scenes.
\newblock In {\em Proceedings of the IEEE Conference on Computer Vision and
  Pattern Recognition}, pages 368--375, 2014.

\bibitem{narayana2013}
M.~Narayana, A.~Hanson, and E.~Learned-Miller.
\newblock Coherent motion segmentation in moving camera videos using optical
  flow orientations.
\newblock In {\em Proceedings of the IEEE International Conference on Computer
  Vision}, pages 1577--1584, 2013.

\bibitem{Brox2014}
P.~Ochs, J.~Malik, and T.~Brox.
\newblock Segmentation of moving objects by long term video analysis.
\newblock {\em IEEE Transactions on Pattern Analysis and Machine Intelligence},
  36(6):1187--1200, June 2014.

\bibitem{patwardhan2008robust}
K.~Patwardhan, G.~Sapiro, and V.~Morellas.
\newblock Robust foreground detection in video using pixel layers.
\newblock {\em IEEE Transactions on Pattern Analysis and Machine Intelligence},
  30(4):746--751, 2008.

\bibitem{Ren2013}
X.~Ren, T.~X. Han, and Z.~He.
\newblock Ensemble video object cut in highly dynamic scenes.
\newblock In {\em Computer Vision and Pattern Recognition (CVPR), 2013 IEEE
  Conference on}, pages 1947--1954, June 2013.

\bibitem{ren2003statistical}
Y.~Ren, C.-S. Chua, and Y.-K. Ho.
\newblock Statistical background modeling for non-stationary camera.
\newblock {\em Pattern Recognition Letters}, 24(1):183--196, 2003.

\bibitem{rowe1996}
S.~Rowe and A.~Blake.
\newblock Statistical mosaics for tracking.
\newblock {\em Image and Vision Computing}, 14(8):549--564, 1996.

\bibitem{Sheikh2009}
Y.~Sheikh, O.~Javed, and T.~Kanade.
\newblock Background subtraction for freely moving cameras.
\newblock In {\em 2009 IEEE 12th International Conference on Computer Vision},
  pages 1219--1225, Sept 2009.

\bibitem{Sheikh2005}
Y.~Sheikh and M.~Shah.
\newblock Bayesian modeling of dynamic scenes for object detection.
\newblock {\em IEEE Transactions on Pattern Analysis and Machine Intelligence},
  27(11):1778--1792, Nov 2005.

\bibitem{shi2000normalized}
J.~Shi and J.~Malik.
\newblock Normalized cuts and image segmentation.
\newblock {\em IEEE Transactions on pattern analysis and machine intelligence},
  22(8):888--905, 2000.

\bibitem{Stauffer1999}
C.~Stauffer and W.~E.~L. Grimson.
\newblock Adaptive background mixture models for real-time tracking.
\newblock In {\em Computer Vision and Pattern Recognition, 1999. IEEE Computer
  Society Conference on.}, volume~2. IEEE, 1999.

\bibitem{Sundaram2010}
N.~Sundaram, T.~Brox, and K.~Keutzer.
\newblock Dense point trajectories by gpu-accelerated large displacement
  optical flow.
\newblock In {\em European conference on computer vision}, pages 438--451.
  Springer, 2010.

\bibitem{Torr2000}
P.~Torr and A.~Zisserman.
\newblock Mlesac.
\newblock {\em Comput. Vis. Image Underst.}, 78(1):138--156, Apr. 2000.

\bibitem{torr2001integrated}
P.~H. Torr, R.~Szeliski, and P.~Anandan.
\newblock An integrated bayesian approach to layer extraction from image
  sequences.
\newblock {\em IEEE Transactions on Pattern Analysis and Machine Intelligence},
  23(3):297--303, 2001.

\bibitem{hopkin2007}
R.~Tron and R.~Vidal.
\newblock A benchmark for the comparison of 3-d motion segmentation algorithms.
\newblock In {\em 2007 IEEE Conference on Computer Vision and Pattern
  Recognition}, pages 1--8, June 2007.

\bibitem{wang1994representing}
J.~Y. Wang and E.~H. Adelson.
\newblock Representing moving images with layers.
\newblock {\em IEEE Transactions on Image Processing}, 3(5):625--638, 1994.

\bibitem{yan2006general}
J.~Yan and M.~Pollefeys.
\newblock A general framework for motion segmentation: Independent,
  articulated, rigid, non-rigid, degenerate and non-degenerate.
\newblock In {\em European conference on computer vision}, pages 94--106.
  Springer, 2006.

\bibitem{Zamalieva2014}
D.~Zamalieva, A.~Yilmaz, and J.~W. Davis.
\newblock A multi-transformational model for background subtraction with moving
  cameras.
\newblock In {\em European Conference on Computer Vision}, pages 803--817.
  Springer, 2014.

\bibitem{Zhu03LabelProp}
X.~Zhu, Z.~Ghahramani, and J.~Lafferty.
\newblock Semi-supervised learning using gaussian fields and harmonic
  functions.
\newblock In {\em IN ICML}, pages 912--919, 2003.

\bibitem{zhu2003semi}
X.~Zhu, Z.~Ghahramani, J.~Lafferty, et~al.
\newblock Semi-supervised learning using gaussian fields and harmonic
  functions.

\end{thebibliography}
}

\end{document}